\documentclass[format=acmsmall]{acmart}

\usepackage{booktabs} 

\usepackage{natbib}
\usepackage{makecell}
\usepackage{bm}

\usepackage[ruled]{algorithm2e} 

\SetAlFnt{\small}
\SetAlCapFnt{\small}
\SetAlCapNameFnt{\small}
\SetAlCapHSkip{0pt}
\IncMargin{-\parindent}

\usepackage{color}

\usepackage{amsmath}

\DeclareMathOperator*{\argmax}{arg\,max}

 \acmDOI{10.1145/3497875}

\received{October 2020}

\begin{document}

\setcopyright{acmcopyright}
\acmJournal{TOIS}
\acmYear{2020} \acmVolume{1} \acmNumber{1} \acmArticle{1} \acmMonth{1} \acmPrice{15.00}


\title[A Re-classification of Information Seeking Tasks and Their Computational Solutions]{A Re-classification of Information Seeking Tasks and Their Computational Solutions}

\author{Zhiwen Tang} \authornote{Both authors contributed equally to the article.} 
\affiliation{%
  \institution{InfoSense, Department of Computer Science, Georgetown University}
  \streetaddress{37 and O Streets Northwest}
  \city{Washington}
  \state{DC}
  \postcode{20057}
  \country{USA}}
\email{zt79@georgetown.edu}

\author{Grace Hui Yang} \authornote{Grace Hui Yang is the corresponding author.} 
\affiliation{%
  \institution{InfoSense, Department of Computer Science, Georgetown University}
  \streetaddress{37 and O Streets Northwest}
  \city{Washington}
  \state{DC}
  \postcode{20057}
  \country{USA}}
\email{huiyang@cs.georgetown.edu}

\begin{abstract}

This article presents a re-classification of information seeking (IS) tasks, concepts, and algorithms. The proposed taxonomy provides new dimensions to look into information seeking tasks and methods. The new dimensions include number of search iterations, search goal types, and procedures to reach these goals. Differences along these dimensions for the information seeking tasks call for suitable computational solutions. The article then reviews machine learning solutions that match each new category. The paper ends with a review of evaluation campaigns for IS systems. 


\end{abstract}

\maketitle


\section{Introduction}\label{sec:introduction}




Information seeking (IS) is a type of human activity where a human user uses a computer search system to look for information. IS is a process and it happens frequently in daily life. For instance, Kids jibber-jabbered with Amazon Alexa, \textit{exploring the limit of its artificial intelligence}. Families search online and gather information from friends and real estate agents to decide \textit{which house to purchase}. Students read textbooks and Wikipedia to understand \textit{what a wormhole} is and \textit{why  Schr{\"o}dinger's cat was both dead and alive}. An IS process is initiated by a human user and  driven through by the user. 
The user, on the other hand, is also driven by some motivation, usually a motivation connected to an information problem. The user carries out iterations of searches to find the needed information. In each search iteration, the user formulates a query, executes the search function, examines the returning results, extracts useful information, and reflects on the results. The iterations go on many cycles until the user's information need is satisfied, or the process is abandoned or no longer supported.  

IS research roots from the Library Science. Early IS work concerns with how a human librarian could serve the patrons to find the desired materials. With the rapid spread of modern search engines, digital libraries, electronic catalogs, and social media, IS research has moved on to making the best use of the new digital tools for the old human-centered task -- assisting human users to seek information. Traditionally, IS shows a much broader scope and higher complexity than the core Information Retrieval (IR) research problem, ad hoc retrieval. The whole or part of IS has been studied under various names in the IR literature. These names include Interactive Information Retrieval (IIR)~\cite{ruthven2008interactive}, Session Search (SS)~\cite{TREC:2014:Session}, Exploration Search (ES)~\cite{white2006supporting}, Dynamic Search (DS)~\cite{DBLP:series/synthesis/2016YangSW}, and Conversational Search~\cite{dalton2020trec}. Plenty of interesting IS research has been generated in the past two decades. 

Researchers have developed taxonomies to describe and organize work in IS and in related research fields. For instance, \citet{marchionini2006exploratory} classified all search activities into three overlapping types -- look-up, learning, and investigation. \textit{Look-ups} are the most basic type of search operations that mainly find facts and answers to a user's query. Typical look-up examples include ad hoc retrieval, Web search, and question answering. Look-ups can be finished in one-shot ad hoc retrieval, or must go through multiple search iterations, each of which fetches an answer to one aspect of a complex information problem. \textit{Learning} is the type of search activities where the user develops new knowledge or skills by reading and comparatively studying the materials returned by the search system. A typical learning example is literature reviewing. \textit{Investigation} is the type of search activities that involve critical assessments of the search results when integrating them into existing knowledge bases and may value recall over precision. Investigations are mostly seen in domain-specific, professional search scenarios, such as legal search, patent prior art search, medical search, and forensics search. 

Later, \citet{white2006supporting} discussed IS with an emphasis on its exploratory nature and sort of renamed the task to ``exploratory search''. The survey covered a wide range of related concepts, including sense-making, information forage, curiosity, and berry-picking, and worked to discern the subtle differences among them. \cite{white2006supporting} also developed a Venn diagram (Figure 3.4 in \cite{white2006supporting}), which organizes many IS-related concepts and tasks in a single graph. The diagram uses a two-dimensional presentation, whose  capacity is too limited to express the complex  relationships that the author would like to express. This is especially true for concepts (categories) with mixed initiatives. In the end, the diagram appears over-crowded and a reader cannot quickly tell apart the concepts.  

More recently, \citet{https://doi.org/10.1002/asi.23617} developed another taxonomy that distinguishes ``look-up'' and ``exploratory search''. The term ``look-up'' in~\cite{https://doi.org/10.1002/asi.23617} has a slightly different meaning from the ``look-up'' used in Marchionini's \cite{marchionini2006exploratory}. In \cite{marchionini2006exploratory}, ``look-up'' has a stronger sense of checking up items from a knowledge base or an inverted index, where the definition puts more emphasis on describing the actions being taken in the process. While in \cite{https://doi.org/10.1002/asi.23617},  ``look-up'' is defined more from perspective of the IS task's end target; and it refers to a process that has discrete, well-structured, and precise search target(s). The latter is similar to one of our new categories. In addition, \citet{https://doi.org/10.1002/asi.23617}'s ``exploratory search'' refer to IS tasks that are imprecise and open-ended, which is also slightly different from the term's initial meaning in White's \cite{white2006supporting}. In \cite{white2006supporting}, ``exploratory search'' almost covers all types of IS tasks; while Athukorala et al. limit the term's use to only refer to tasks that have a strong exploratory and open-ended nature. This is similar to another new category we present in this paper. 


We observe that these existing taxonomies mainly derive their categories from the user-side and are mostly descriptive. While they are very useful for user-side studies, when developing IS systems, however, these imprecise and descriptive categorizations make it difficult for practitioners to design computational solutions and program them. We realize that an easy matching between IS tasks and suitable computational models is missing. This may explains that despite the advancement of the user-side studies, the system-side study of IS remains underdeveloped.

In this article, we propose a re-classification of IS tasks and methods, aiming to reveal the computational structure embedded in the IS tasks. The new taxonomy classifies IS works along a few critical dimensions, which serve for understanding both the tasks and the computational methods. The proposed dimensions are
\begin{itemize}
    \item \textit {number of search iterations},
    \item  \textit {search goal types}, and 
    \item  \textit{procedures to reach these goals}.
\end{itemize}
We also review the literature and classify the works using the new taxonomy. This practice of reviewing the old works from new angles, which have a heavy flavor of computational thinking, allows us to develop new understanding of the field. Our goal is to enable easier use of effective computational tactics, achieving full automation of IS in the era of Artificial Intelligence (AI).  

\begin{figure}[h]
    \centering
    \includegraphics[width=0.5\columnwidth]{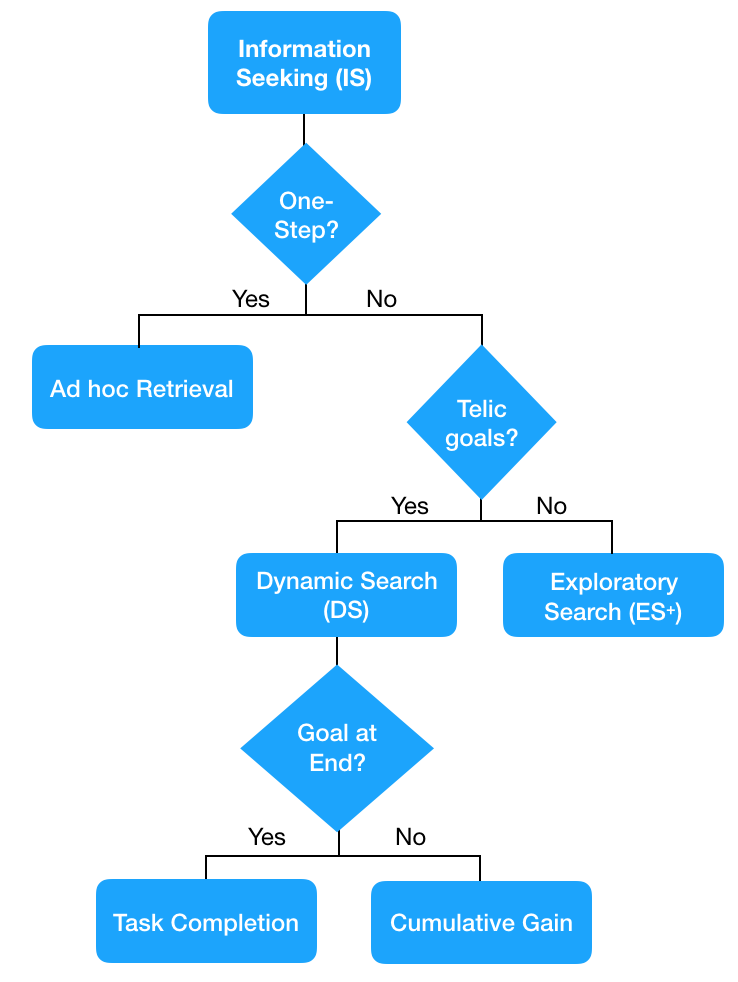}
    \caption{Re-classification of Information Seeking (IS) Tasks. This figure shows our new taxonomy for the IS tasks. The classification is done based on three dimensions, which need to be applied in order. First, we decide if the search task is a single turn or multiple turn task. If it is single turn, it belongs to the well-known ad hoc retrieval. Second, if it is multi-turn, we further examine the search task's goal type based on if the task's motivation is serious and success-driven. The tasks then map into Dynamic Search and Exploratory Search. Note here our use of the term Exploratory Search is slightly different from \cite{white2006supporting}, so its acronym is ES+. Third, for Dynamic Search, we further distinguish if the search goal is achieved only at the end of the process or can be accumulated part by part throughout the process. They then correspond to the Task Completion and Cumulative Gain categories, respectively. }
    \label{fig:rough-relation}
\end{figure} 

Figure \ref{fig:rough-relation} illustrates our taxonomy.~
The first dimension is \textit{the number of search iterations} involved in the process. A search iteration is a complete cycle of (1) user issuing a query, (2) search engine retrieving documents for the query, and (3) the user consumes the returned documents and evaluates if they meet the information need. This dimension is an evident difference between ad hoc retrieval and the broader IS process: Ad hoc retrieval runs only one iteration while IS runs multiple. Ad hoc retrieval is mainly responsible for \textit{look-ups} in~\cite{marchionini2006exploratory}'s classification. On the contrary, IS takes numerous iterations to finish \textit{learning} or \textit{investigation} tasks. As the number of search iterations increases in an IS process, queries change from iteration to iteration. The user's understanding of the information needs also evolve. Multi-turn Question Answering (QA), conversational search/AI, dialogue systems, and dynamic search fall under this category. 

The second dimension in our taxonomy is task goals related to user motivations. The motivation for a human user to start an IS process can vary. 
In psychology,  they categorize people's motivations in everyday life along a few dimensions~\cite{personalitydynamics}. They include ``means-ends'', ``rules'', ``transactions'', and ``relationships''. 
The most significant dimension, ``means-ends'', is about achievements, goals, and enjoyments of process. It has two opposing states. They are  {\it telic}, at which one is motivated by achievement, task completion, and fulfilling of goals; and {\it paratelic}, at which one is playful and seeks excitement and fun. 
We propose to use the ``means-ends'' states to classify further IS tasks into {\it tasks with telic goals} and {\it tasks with paratelic goals}. Tasks with telic goals are also known as task-oriented. We can link them to more  ``serious'' search tasks such as finance, health care, legal issues, and family affairs. These more ``serious'' search tasks are closely related to one's need for problem-solving, decision making, and reasoning. 
We borrow the term ``dynamic search'' (DS) from \citet{DBLP:series/synthesis/2016YangSW,DBLP:conf/sigir/YangS014} to call this category. 

On the other hand, tasks with paratelic goals are primarily driven by curiosity. They appear to be more open-ended. These tasks' goals are ``to look for new insights,'' \cite{white2006supporting}, ``being exploratory'', or ``to be entertained''. We use the term ``exploratory search'' from \citet{https://doi.org/10.1002/asi.23617} to name this category. Note this category's scope is narrower than the word initially means in \cite{white2006supporting}. We think in \cite{white2006supporting}, the term is used to cover a scope bigger than dictionary meaning can cover. In \cite{white2006supporting}, ``exploratory search'' (ES) seems to be used to refer to all types of IS tasks. However, in a dictionary, the word ``exploratory'' only means to be related to exploration. In this work, we would like to restore its scope suggested by the dictionary and only refers to tasks with paratelic goals. Because of this subtle change, we name this category ``ES+'' to show that it is a variation from the more known ES used in \cite{white2006supporting}. 

The third dimension is a further division for IS tasks with telic goals based on how to arrive at their goals. We classify them by whether these tasks have an ending goal state or not.  The first type of goals can be achieved by a workflow of conditions, including preparations, planning, developments, casual dependency, completing subtasks, and finally reaching the goal. The second type can achieve their tasks by collecting (accumulating) gains over a series of subtasks without any ending goal. ``The more, the merrier'' is the belief. We call these two types ``task completion'' and ``cumulative gain,'' respectively.

The remainder of the paper is organized as the following. Next, we detail the second and the third dimensions in Sections \ref{telic} and \ref{match}. The following sections review existing computational IS solutions based on our taxonomy. Sections~\ref{iir}, \ref{bandits}, \ref{rl:value-based}, \ref{imitation}, and \ref{open-ended} present local methods, multi-armed bandits (MAB), value-based RL, policy-based RL,  model-based RL, supervised imitation learning, and open-ended RL, respectively. Finally, Section \ref{eval} reviews a few evaluation campaigns for IS systems.

\section{Re-classification by Task Goals} 
\label{telic}

\begin{table}[t]
\centering\small
\begin{tabular}{ p{4cm}|p{4cm}}
\hline
Telic & Paratelic \\
\hline
$\bullet$ \textit{\small{Serious. Focus on future goals and achievement. Tend to avoid arousal, risk \& anxiety.}} & $\bullet$ \textit{\small{Playful, passion and fun. Focus on current moment. Seek excitement and entertainment.}} \\
\hline
\end{tabular} \caption{Goal Types in the Means-Ends Domain. The two goal types that we use in this paper are inspired from psychology studies on human motivations. Psychology studies group human motivations into four domains, including ``means-ends,'' ``rules,'' ``transactions,'' and ``relationships''. Here we propose to use the first domain ``means-ends", which is about achievements, goals, and enjoyments of process. It concerns two motivational states, {\it telic} -- at which one is motivated by achievement, task completion, and fulfilling of goals-- and {\it paratelic}--at which one is playful and seeks excitement and fun. A telic example is that people run because they want to win a medal. When doing the same activity, running, when people have paratelic goals, they run because they enjoy running itself, not because they want to win a medal~\cite{personalitydynamics}. In the context of IS, telic and paratelic states parallel to goal-oriented and non-goal-oriented search processes. We name them Dynamic Search and Exploratory Search in our taxonomy in Figure \ref{fig:rough-relation}.} \label{tab-reversal-theory}
\end{table} 

IS is the umbrella research topic we discuss in this paper. There is always a motivation to start an IS process. Therefore it is essential to understand the motivation types. Psychology studies group human motivations into four ``domains.'' These are ``means-ends,'' ``rules,'' ``transactions,'' and ``relationships'' \cite{michael1989reversal}. Table \ref{tab-reversal-theory} show the first two domains. Each dimension consists of a pair of opposing motivational states. At one moment, a person can only be at one of the two states. The two opposing motivational states in one domain ``reflect people's motivational styles, the meaning they attach to a given situation at a given time, and the emotion they experience''~\cite{personalitydynamics}. 

The first domain, ``means-ends'', is about achievements, goals, and enjoyments of process. Its two states are  {\it telic}, at which one is motivated by achievement, task completion, and fulfilling of goals; and {\it paratelic}, at which one is playful and seeks excitement and fun. When one is telic, she is serious about the task at hand and focuses on achieving the task's goal. For instance, people run because they want to win a medal. Whereas when one is paratelic, the activity she does is not for the sake of the task's goal but the task's own sake. For instance, people run because they enjoy running itself, not because they want to win a medal~\cite{personalitydynamics}. In the context of IS, telic and paratelic states parallel to goal-oriented and non-goal-oriented search processes. We call them dynamic search (DS) and exploratory search (ES+) in this paper. 
A user at the telic state would focus on finishing a search task, for instance, to look for a job vacancy or medical help. In contrast, a user at the paratelic state seeks to enjoy the search process itself, browsing for funny videos on YouTube. 

We can identify these task goal types by analyzing query logs. Modern commercial search engines record user activities and keep search logs that contain rich interaction signals such as queries, dwelling time, and clicks. However,  boundaries between search sessions, which overlap with individual IS tasks,  are not explicitly marked in the logs. Much research has been devoted to segment search sessions or identifies search tasks from the logs.  \citet{DBLP:conf/cikm/JonesK08} discovered that it was of limited utility to claim a separate session after a fixed inactivity period since many sessions could be interleaved and hierarchically organized. Instead, they recognized individual search tasks via supervised learning, with temporal and word editing features. 
\citet{DBLP:conf/www/WangSCHWC13} worked on long-term search task recognition with a semi-supervised learning method that exploited inter-query dependencies derived from behavioral data. Work by \citet{DBLP:conf/wsdm/LuccheseOPST11} utilized Wiktionary and Wikipedia to detect semantically related query pairs. However, some still found that a correctly set threshold of inactivity period would work adequately to separate sessions, for instance, a 30-minute inactive threshold. We can use this method in a wide range of online applications \cite{DBLP:conf/www/HalfakerKKTNSUW15}.

\subsection{Telic Search Tasks - Dynamic Search (DS)}

Telic search tasks, aka DS, are IS tasks motivated by achievement, task completion, and fulfilling purposes. We can see everyday search tasks with telic goals in conversational search, dynamic search, and task bots.  \citet{DBLP:conf/chiir/RadlinskiC17} proposed a  general framework for conversational search without assuming a high task formality. They define conversational search as a mixed-initiative communication running back-and-forth between a user and a dialogue agent, without distinguishing task bots from chatbots. When generating conversations, the agent focuses more on eliciting user preferences and identifying the user's search target. Their work laid out a blueprint for a wide range of research domains that a conversational search system would involve. 

Telic search tasks assume a highly structured task formality.  We can use a metaphor to describe this. For instance, ``way-finding''. It emphasizes that the user must conceptualize the sought-after information first before navigating to the desired part.
For another example, to {\it makes a doctor's appointment}, a task bot would need to know the \textit{when}, \textit{where}, and \textit{who} of the visit; to {\it finds a target smart phone}, the task bot may ask the \textit{brand}, \textit{price}, \textit{color}, and \textit{storage capacity}. Because of this high task formality, simple techniques such as slot-filling appear to work well.

Many IS evaluations are for tasks with telic goals-for instance, the TREC Dynamic Domain (DD) Tracks \cite{DBLP:conf/trec/YangTS17,DBLP:conf/trec/YangS16,DBLP:conf/trec/YangFS15} and the TREC Session Tracks \cite{TREC:2010:Session,TREC:2011:Session,TREC:2012:Session,TREC:2013:Session,TREC:2014:Session,DBLP:conf/sigir/CarteretteCHKS16}. 
Recently, led by the Amazon Alexa Taskbot Challenge~\cite{DBLP:journals/corr/abs-1801-03604}\footnote{https://developer.amazon.com/alexaprize/challenges/current-challenge/taskbot/faq}, research in task bot has received much attention. A standard pipeline used in a task bot includes the following.  First, human utterances go through a natural language understanding (NLU) component where the task bot parses and interprets the voice input. The task bot then tracks dialogue states by methods such as {\it slot filling}. Second, a ``policy learning'' module learns from the converted training data to decide a dialogue act based on the state. The module can use expert-crafted rules~\cite{DBLP:journals/cacm/Weizenbaum66}, supervised machine learning~\cite{DBLP:conf/aaai/SerbanSBCP16}, or reinforcement learning~\cite{DBLP:conf/emnlp/LiMRJGG16}. 
Third, a natural language generation (NLG) component generates the task bot's responsive utterances from the dialogue acts determined by the policy. 

We train modern task bots end-to-end with deep supervised learning. Based on how they generate/select their response,  we can categorize these systems as retrieval-based or generation-based. Retrieval-based task bots select the reactions from a pre-built candidate index, choosing the one that suits the context best with text retrieval methods \cite{Yang:2018:RRD:3209978.3210011,DBLP:conf/iclr/BordesBW17,DBLP:journals/corr/abs-1712-07229,DBLP:conf/acl/WuWXZL17,DBLP:conf/sigir/YanSW16,DBLP:conf/emnlp/ZhouDWZYTLY16,DBLP:conf/cikm/ZhangCA0C18,DBLP:conf/acl/QiuYJZHCCL18}. Generation-based task bots generate the responses with NLG techniques \cite{DBLP:conf/acl/LiGBSGD16,DBLP:conf/acl/ShangLL15,DBLP:conf/naacl/SordoniGABJMNGD15,DBLP:conf/acl/TianYMSFZ17,DBLP:journals/corr/VinyalsL15,DBLP:conf/cikm/PatidarAVS18}. Generation-based approaches appear to be more flexible but also increase the complexity of the system.


\subsection{Paratelic Search Tasks - Exploratory Search (ES+)}


 

Paratelic search tasks, aka ES+, are IS tasks essentially triggered by curiosity. These activities are not motivated by a telic goal. Instead, human's desire to understand and to learn is the pure drive. The users perform the search tasks only for the tasks' own sake. They enjoy searching. ES+ often happens when users face a new domain or a domain of interest. 

In~\cite{white2006supporting}, it said ES might eventually yield knowledge-intensive products such as a research paper or a good understanding of the housing market. I also said ES typically involves high-level intellectual activities such as synthesis and evaluation. In this paper, our ES+ does not include these meanings. Because the above suggests a telic goal and brings ambiguity in ES, IS, and other IS types. 

Our ES+ demonstrates a few unique features. First, ES+ has a strong sense of being ``exploratory''. ``Berry-picking'' is a metaphor given by~\citet{bates1989design}. It refers to the evolution of an information need. A user starts with a vague perception of an information need and traverses the document collection, collecting information along the way. Newly-collected fragments may change the information need and the user's behavior for the next round of ``picking.'' Second, ES+ is open-ended. Its users tend to spend more time on the task.  Even after they have gathered all the information fragments they need, they may continue searching for the own sake of searching. The users tend to keep learning and validating the information from various sources without a telic goal ahead. This behavior is distinct from an IS task with a telic goal. In that, after reaching the goal, the search would be considered complete and end. Third, users in ES+ could be unsure about what they are looking for and how they can achieve their non-telic goals. As a result, on top of issuing queries and examining results, ES+ users may also need more time to synthesize search results into their knowledge base. 
Chatbots \cite{DBLP:journals/corr/JiLL14,INR-074} have received tremendous attention in recent years. For example, the Amazon Alexa Prize has hold competitions for three years on social bots.\footnote{https://developer.amazon.com/alexaprize/challenges/current-challenge/} How they differ from text-based IS is probably only in the media of interaction. Text-based IS uses documents to communicate to users, and chatbots use machine-generated natural language (NL) utterances. The use of voice input marks the only difference between the two kinds. In social bots, the system aims to have a long engagement time with the user. And the user uses social bots for chit-chat, discussion, or explore of exciting topics.

\section{Re-classification by Goal-Reaching Procedures}

\label{match}


\begin{table*}[t]
    \centering  \small
    \begin{tabular}{l|c|c|c}  
    \hline 
     &   \multicolumn{2}{c|}{IS w/ Telic Goals} & IS w/ Paratelic Goals \\ 
    \hline
            & {\bf Being cumulative} & {\bf Being successful} & {\bf Being enjoyable} \\ 
    \hline 
    Example  & Dynamic Search & Dynamic Search & \\ 
    Research  & Recommender Systems & Taskbots  & Exploratory Search \\
     Problems      & eDiscovery &  Task-based Dialogue Systems & Social bots \\
           & Online L2R & Conversational AI/Search & \\ 
           & & Multi-turn QA & \\
    \hline
    Task Illustration  &  &  &    \\ 
    &  
   \includegraphics[width=0.18\textwidth]{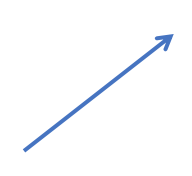}
   \label{fig:gain} 
  &   \includegraphics[width=0.18\textwidth]{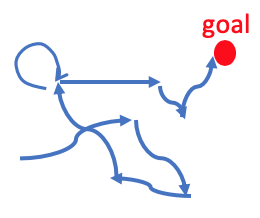}
   \label{fig:success} &   \includegraphics[width=0.18\textwidth]{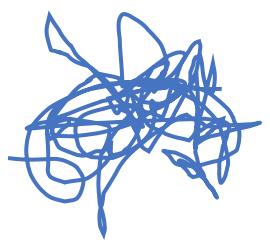}
   \label{fig:enjoyable} \\ 
    \hline 
         Gain Curves &  &  &    \\ 
    &  
   \includegraphics[width=0.18\textwidth]{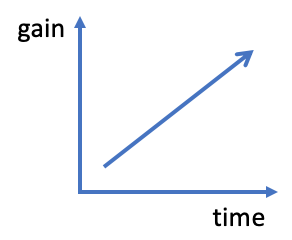}
   \label{fig:gaingain} 
  &   \includegraphics[width=0.18\textwidth]{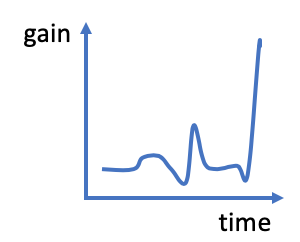}
   \label{fig:successgain} &   \includegraphics[width=0.18\textwidth]{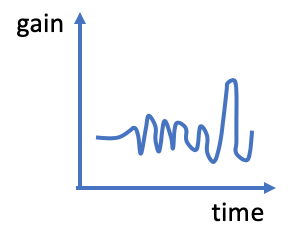}
   \label{fig:enjoyablegain} \\ 
    \hline 
   Suitable ML  & Bandits Algorithms & Policy-Based RL & Value-Based RL \\
   Methods& Local Methods & Actor-Critic & Model-Based RL \\
         & & Model-Based RL & Open-Ended RL \\
         & &  & Local Methods \\
   \cline{2-4}
   & \multicolumn{3}{c}{Imitation Learning} \\
    \hline
    \end{tabular}
    \caption{Gain vs. Success vs. Entertainingness. This table's columns 2 to 4 show the three types of goal-reaching procedures of IS tasks. The procedures are aiming for being cumulative, successful, and enjoyable. They correspond to Figure 1's taxonomy categories ``Task Completion'', ``Cumulative Gain'', and ``Exploratory Search+'', respectively. Their optimization goals are different too. They aim to optimize over gains, success, and entertainingness. These imply very different optimization goals for an algorithm and IS system. We also draw task illustrations of how a human's search trajectory should look like. In addition, we sketch typical gain curves over time if the cumulative gains are plotted. For the success-driven and entertainment-driven tasks, there might even be reduction of gains at certain points of time. Based on the tasks' nature, we also list some suggestions to suitable family of ML solutions. For instance, we think for cumulative IS tasks, Bandits algorithms and Local IR feedback methods would work well. For success-driven (aka task-completion-oriented) tasks, policy-based RL, Actor-Critic, and Model-Based RL would be good choices. Each type will be discussed in the later sections.} 
    \label{tab:goal_ending}
\end{table*}

The third dimension in our taxonomy is the procedure to reach its goal within an IS task. Given the different goal types we have previously discussed, we take separate procedures to achieve telic goals or paratelic goals. In fact, under the same category of telic goals, we can further divide the tasks into ``gain-based'' telic tasks and ``success-based'' telic tasks. The difference lies in how the end goal in the task is placed and reached. In the ``gain-based'' telic tasks, there is no end goal. The goal of the entire search process is to get more and more gains. As time goes by, it will have a monotonically increasing cumulative gain. In the ``success-based'' telic tasks, there is an ending goal state to signal the success and completion of the task. Before reaching the goal condition, any steps could receive little reward or gain. However, it does not mean they are not helpful steps to lead to the final success. Therefore the gain curve for the ``success-based'' telic tasks shows a big jump at the end, while ``gain-based'' expects steady growth with no upper bound. 

Table \ref{tab:goal_ending} shows our idea on the third ``goal-reaching procedure'' dimension. First, we list a few example research problems for each type. Second, we draw illustrations of how the tasks or search paths would be. Third, we also draw expected gain/relevance vs. time curves. Fourth, we recommend suitable families of ML solutions for these different types of IS tasks. Our recommendation is made based on the tasks' nature. By how to reach their task goal during an IS process, we group IS tasks into three types: {\it being cumulative}, {\it being successful}, and {\it being enjoyable}. The first two types are telic tasks, while the third is paratelic tasks. 

The first type involves a process that collects relevant information along the way and maintains a monotonically increasing cumulative gain. At any point of the process, positive gains can be collected and added to the overall metric to maximize. These IS tasks do not require algorithmic convergence to an (optimized) value nor to end with success. It has no concern over being successful or not. The tasks are more about collecting gains. Example applications include recommender systems, online L2R,  eDiscovery, and those DS tasks with cumulative gain as the goal.\footnote{Most search tasks in the TREC DD, TREC Session, and TREC CaST belong to this category.} The most suitable ML methods include (1) bandits algorithms due to their stateless status and no specific requirement to `win' an overall game, and (2) most early interactive IR methods created in the IR community.  

The second type include IS processes that aim for  ``success'' to win at the end of their tasks. The goal can be \textit{successfully} satisfying a user's information need (only if the need is to be successful), finishing a task, closing a deal, making a transaction, or winning a game. The task environment gives a big final reward when the agent succeeds in the end. These IS tasks go through stateful task steps, converge to an optimum and end with success. Example applications include task bots, conversational AI/search, task-based dialogue systems, and multi-turn QA and dynamic search when they have a task to complete. 
For these stateful problems, policy-based RL is the most suitable family of ML methods, given its emphasis on gradual optimization towards success. It implies RL can bear with strategic loss in the short term. Besides, an agent can have multiple winning strategies by identifying multiple search paths to reach the end goal. Similarly, actor-critic methods, which share the same gradient climbing towards success with policy-based RL, are also plausible. Model-based RL, used in combination with policy-based RL in its policy learning part, is suitable.


The third type of goal-reaching procedure is the IS tasks that have the purpose of being ``being enjoyable''. It is closely linked to being engaged,  entertained, curious,  or inspired. This category corresponds to IS tasks with paratelic goals. The search paths can be pretty random, and the gain cures can have up-and-downs and never converge. Example applications include exploratory search and social chatbots. 
Exploration led to many discoveries in human history. The human drive of having fun is perhaps as crucial as ``being cumulative'' and ``being successful.'' Suitable ML methods for this type can be value-based RL \cite{DBLP:books/lib/SuttonB98}, model-based RL with value-based policy learning, open-ended RL \cite{wang2019paired,wang2020enhanced} and early local methods from the IR community. These methods are open, only accessible to a local area in the search space, without knowing a global optimal point.

All above goal-reaching procedures are sequential decision-making problems. Therefore,  supervised imitation learning applies to them all when training demonstration is available. 

We survey these ML solutions to IS and organize them by our new taxonomy in the following sections. We present each family of solutions in rough order from the least amount of global optimization and planning involved to the most (except open-ended RL due to its underdevelopment). They are (1) early IIR methods, such as relevance feedback and active learning. We call them ``local methods'' as they only have access to local search spaces and decide subsequent retrieval actions accordingly. (2) Bandits algorithms commonly used stateless methods in recommender systems and online L2R. (3) Value-based RL methods. (4) policy-based RL methods and actor-critic. (5) Model-based RL that combines inference and policy learning. (6) Supervised imitation learning. And (7) open-ended RL is still ongoing research but essential and relevant in our context. 
Note that our re-classification may make our view quite different from the usual opinions to the pieces of reviewed work.

\section{Local Methods}
\label{iir}


Early methods in IR dealing with interactive or IS are named Interactive Information Retrieval (IIR)~\cite{ruthven2008interactive}. Here we call them ``local methods,'' for they have no access to global search space, which distinguishes them from other methods from ML. They are perhaps the most ``native'' methods from the IR community and widely used in IR applications.

These local methods run with multiple search iterations, where the focus is on how to adjust the ranking function based on user feedback. These methods assume the existence of relevant documents in the collection. They iterative the search by either reformulating queries or modifying the search space based on feedback. However, local does not consider the internal structure (or statefulness) of the information need. They use expert-crafted rules, instead of data-driven models, to guide the exploration and results diversification in the solution space.

{\bf Relevance Feedback.} Relevance feedback (RF)  \cite{DBLP:conf/sigir/LavrenkoC01,DBLP:journals/jasis/RobertsonJ76} is perhaps the most well-known local method. After observing results retrieved by a search engine, the user provides feedback to the search engine. The feedback indicates which documents are relevant and which are not. The search engine then makes modifications in its algorithm to bias what the feedback prefers. The change can only affect the next run of retrieval, either in the query space or the document space. 

Rocchio is a representative RF algorithm \cite{rocchio1971relevance}. Given user feedback that indicates relevance (+ve) and irrelevance (-ve) for returning documents, Rocchio modifies the original query representation vector  $q$ to $q'$, moving $q'$  closer to the feedback points:
\begin{equation}
    \bm{V}(q') = a \bm{V}(q)  + b \frac{1}{|\mathcal{D}^+|} \sum_{d_i \in \mathcal{D}^+ } \bm{V}(d_i) - c \frac{1}{|\mathcal{D}^-|}  \sum_{d_j \in \mathcal{D}^-} \bm{V}(d_j)\,,
\end{equation}
where $\bm{V}(*)$ is the vector representation for a query. It can take the form of bag-of-words (BoW). Here $q$ is the initial query, and $q'$ is the modified query. User feedback is encapsulated in $\mathcal{D}^+$ for relevant documents  and $\mathcal{D}^-$ for irrelevant documents.  Scalars $a, b$, and $c$ are coefficients that We can tune. 

Rocchio attempts to improve retrieval for the same query $q$. The modified $q'$ is like a ``mock query''. But it is not a different query that carries forward the information seeking task. The feedback is to improve retrieval for $q$. Although not originally designed for a sequence of queries, RF models' idea of modifying the query (or document) space has much-inspired research in DS systems. Direct uses can be found in \cite{jiang2013pitt,DBLP:conf/sigir/LevineRC17,DBLP:conf/sigir/ShalabyZ18}. \citet{jiang2013pitt} combined the current search query, past search queries, and clicked documents to update the document-to-query relevance score, with the duplicated documents demoted. \citet{DBLP:conf/sigir/LevineRC17} exploited query term changes (difference between $q_t$ and $q_{t-1}$), which reflect the state transition $M$, to adjust term weights for both queries and documents. \citet{DBLP:conf/sigir/ShalabyZ18} re-ranked documents in a neural L2R model by extending its word2vec \cite{DBLP:conf/nips/MikolovSCCD13} representation with documents in the positive feedback. 

Unlike the RF methods, other local approaches do not directly calculate document relevance scores.  Instead, they focus on reformulating the queries (not the query space) and then feeding the new queries into an existing ad hoc retrieval function. There are three major approaches to query reformulation: (1) {\it query expansion}, (2) {\it query re-weighting}, and (3) {\it query suggestion.}     

{\it Query expansion} adds more terms into the original query.  Many works extend the current query based on past queries in the same session \cite{liu2011dutir,yuan2012u,DBLP:conf/wsdm/WangBHZ18,adeyanju2011rgu}.  \citet{liu2011dutir} augmented the existing query with all past queries within the same search session, with heavier weights on the more recent queries. \citet{yuan2012u} also excluded terms predicted to be less effective. Queries can expand with snippets \cite{matthias2013webis}, anchor text \cite{kruschwitz2012university},  clicked documents \cite{carterette2011implicit}, or pseudo-relevance feedback \cite{king2010cengage}. \citet{bah2014university} expanded queries based on queries suggested by a commercial search engine.  

{\it Query re-weighting} does not change the membership of terms in a query. It only adjusts the weights of each query term. In this sense, it is similar to what we have seen in Rocchio's query space modification. \citet{zhang2012bupt_pris} manually crafted rules to re-weight query terms given the user's attention span and clicks. \citet{liu2012rutgers} classified search sessions into four types and assigned a unique set of pre-defined parameters to re-weight query terms. The four session types are determined based on (1) whether the search goal is specific or amorphous and (2) whether the expected outcome is factual or intellectual.

{\it Query suggestion} is mainly done with the help of external resources. Query logs are popular external resources for this purpose. When query logs from other users are available, We can use them to suggest new queries to help the current user. The assumption is similar search tasks would use similar queries, even by different users. 
Click graph analysis can recognize patterns in massive user data and then suggest queries. A click graph  \citet{DBLP:conf/sigir/CraswellS07} is a bipartite graph between queries and clicked documents. An edge between a query and a document indicates a user has clicked on the document for that query. We can estimate the edge weights via random walk. For instance,
\citet{feild2014endicott,he2011cwi} uses click graphs to induce new queries, and \citet{DBLP:conf/cikm/OzertemVL11} creates a set of suggested queries that aim to maximize the overall query utilities. We can also make query suggestions with the help of external knowledge base \citet{broccolo2012generating}. 

{\bf Active Learning.} Relevance feedback is sometimes used in combination with active learning. Active learning is a type of semi-supervised learning that allows the learning agent to actively prompt a human expert for labels (feedback) when needed. The main challenge for the learner is to select which data samples to ask for a label. This selection needs to be smart because human labeling is costly. A common practice is to select those who fall near the decision boundaries. \citet{DBLP:conf/sigir/TianL11} proposed using active learning for IS. They use a support vector machine (SVM) to score the documents' relevance. At each search iteration, maximally uncertain documents -- i.e., documents that are the closest to the SVM decision boundary -- are selected and shown to the user. We expect the user to give feedback about these documents' relevance. \citet{DBLP:conf/sigir/LiuBC12} predicted a document's usefulness via a decision tree (DT). The tree grows by recursively adding the most indicative user behavior features, such as length of dwelling time or the number of visits, from the relevance feedback. The tree then estimates the use of documents and to modify the query space. 

It is worth noting that active learning approaches would interrupt the natural communication flow between user and search engine because the search engine would need to stop and ask questions. The search engine agent is a lot more `active' here.  It turns out the entire information seeking process is led more by the search engine, not by the user.

Besides the above, local methods have reached out to exciting fields. For instance, \citet{DBLP:conf/sigir/Azzopardi11} studied economic models for IS. The work modeled information seeking as generating value (relevance) by spending costs ( writing queries and assessing with feedback). With microeconomics theory applied, a good search strategy should minimize the cost by minimizing the number of queries and assessments needed in the process. This line of research well explained several user search behaviors with the economics models \cite{DBLP:conf/sigir/Azzopardi14} in their follow-up work. \citet{albakour2011university} proposed a local model inspired by the human immune system \cite{DBLP:journals/nc/NanasR09}. They represented an information need as a network of terms. In the network, each node is a word, and each edge is the words' affinity. Words are then ordered by decreasing weights, and they form a hierarchy. A directed, dynamic spreading model estimates the relevance of documents. The spreading starts by activating matched terms then disseminates part of the energy to the associate terms. 
In this process, (1) words (nodes) matching the information need are activated; (2) their higher-ranked neighbors are also activated by receiving parts of their weights passed down; (3) and these higher-ranked neighbors can continue disseminating the weights to their neighbors. The whole activation process runs until They can reach no nodes. The term weights after this spread calculate the final document relevance score. When they found new documents, they would change the network topology by adding new words and updating the links. This topology change would impact the activation spread process and yield different document relevance scores.  

Although with a low level of automation, these local methods can be pretty efficient. For gain-cumulative search tasks and paratelic search tasks, where good planning of steps is not the concern, they can be moderately effective. However, without global optimization, We can stick them in local minimums. Errors carried from prior search iterations can quickly propagate and yield poor results. They cannot also plan for future retrievals and reach long-term goals.


\section{Bandits Algorithms} 
\label{bandits}

{\bf RL Background.} Reinforcement learning (RL) \cite{DBLP:books/lib/SuttonB98} is a type of machine learning where a learning agent interacts with environments to get feedback and gradually form an optimal decision-making policy. RL originated from \textit{animal learning}, which concerns how animals set up their responses based on observations of the surrounding environments to maximize chances of survival and fertility. RL has many great successes in robotics, computer vision, natural language processing, and finance. 

In RL~\cite{DBLP:books/lib/SuttonB98}, a learning {\it agent} interacts with an {\it environment}, which represents a dynamic world. The agent continuously acts in the environment to figure out a strategy to survive, succeed, or win. Either accumulating adequate rewards in the process or achieving a high-defining final reward can declare the winning. Rewards are feedback provided by the environment to the agent's actions. The sum of the rewards is called {\it return}.  The surviving strategy, also known as the {\it policy}, is a function that maximizes the agent's long-term return when deciding which actions to take in various situations. An RL agent's learning objective is to find the best decision-making policy. 

RL uses a set of symbols to describe its settings. They are $\langle S, A, R, M\rangle$:
\begin{itemize}
    \item $S$:  the set of states.  A state $s_t=s$ is a representation of the environment at time $t$;
    \item $A$:  the set of actions. An action $a_t$ is an operation that an agent can choose to change the state at time $t$;
    \item $R$: the reward function. $R(s_t=s, a_t=a)$ denotes the reward signal given by the environment if the agent takes action $a$ at state $s$.  Its scalar value, sometimes called {\it reinforce}, is denoted by $r$. It can be a function, too, such that $R: S \times A \rightarrow \mathcal{R}$. 
    \item $M$:  the transition function between states.  $M(s_{t+1}=s' | s_t=s, a_t = a)$ is the probability that the environment goes from state $s$ to next state $s'$ if the agent takes action $a$.  
\end{itemize}
 
\begin{figure}[h]
    \centering
    \includegraphics[width=0.45\linewidth]{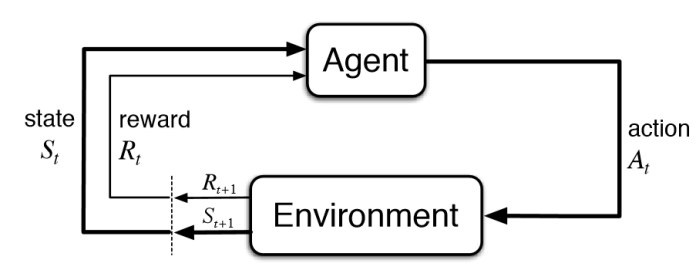}
    \caption{Reinforcement learning (RL) framework. In RL, the learning Agent learns via its interaction experience with an Environment (the world). It is a lot like how a baby learns from scratch. In an environment or closed world, the baby learns her behavior by making mistakes and learn what can be done and what shouldn't. The training and learning goes on until the baby grows up and is mature; then the learning agent can finish the learning and goes to the new world for test. In this trial-and-error setting, the learning Agent takes actions $A_t$ at time $t$ for state $S_t$. The Environment outputs a Reward $R_t$ for this action. The Agent receives $R_t$ and adapts its policy and moves on to handle next state $S_{t+1}$ with a new action. The loop goes on and gradually the agent accumulates sufficient experiences by making all kinds of mistakes, learns from them, and becomes smarter and smarter, until it is confident about the best policy that shows which action to take under what state. Then the learning can be concluded. Figure from \citet{DBLP:books/lib/SuttonB98}.}
    \label{fig:rl}
\end{figure}

These elements work together dynamically. 
At time step $t$, the agent observes the state ($s_t$) from the environment and takes an action ($a_t$). The action impacts the environment, brings in a reward ($r_{t+1}$), and  lands on a new state ($s_{t+1}$). This process loops  until reaching the end, generating a \textit{trajectory}  $s_t,a_t,r_{t+1},s_{t+1},a_{t+1},r_{t+2},..., s_{T},a_{T},r_{T}$. Figure \ref{fig:rl} illustrates a general RL framework \cite{DBLP:books/lib/SuttonB98}.

The learning objective of RL is to learn an optimal policy that helps the eventual winning when responding to various situations. With this policy, the expected return are maximized. A policy is denoted by $\pi(s)$ and is a stochastic function from $s_t$ to $a_t$. The optimization can be expressed by: 
\begin{equation}
    \pi^* = \argmax_{\pi:s\rightarrow a} \mathbb{E}_{\pi:s\rightarrow a, t=1,...T} G_t\,,
\end{equation}
where return $G_t$ is the discounted cumulative rewards:
\begin{equation}
    G_t =\sum_{k=0}^{\infty}\gamma^k R_{t+k+1} (s_{t+k+1}, a_{t+k+1})\,,
\end{equation}
where $\gamma \in [0,1]$ is a factor discounting future rewards. Thus, the expected return is the expected long-term rewards over the entire course of interactions, such as one complete game or one accomplished task.

RL agent's learning has two traits, {\it adaptive} and {\it exploratory}. Learning and optimization are about being the best adaptor in a dynamic environment and handling issues emerging in diverse situations. The agent may have encountered some problems during the training, and others may be unexpected. This ability to be adaptive allows RL agents to handle unseen challenges better. Being exploratory is another trait. RL permanently recognizes the value of exploration. From the outset of the field, not being greedy has been a build-in functionality in RL.    
    
In the approaches that we review in this paper, most of them share a common setting. A search engine is the RL agent, and the environment is either (a) the ensemble of user and document collection or (b) document collection alone. Option (b) models the user as another agent. In the more commonly used (a), the search engine agent observes the environment's state ($\bm{s}_t$) at time $t$,  takes actions ($a_t$) to retrieve documents and shows them to the user. The user provides feedback, $r_t$, which expresses the extent to which the retrieved documents satisfy the informational need. The documents retrieved (deduced) by the search engine agent's action would probably also change the user's perceptions about what is relevant and what stage the search task has progressed. These changes are expressed as state transitions $\bm{s}_t \rightarrow \bm{s}_{t+1}$. The optimization goal is to maximize the expected long-term return for the entire information seeking task. 
    
A common misconception is that an RL agent gets trained and produces results at the same time. We suspect this comes from an ambiguity between RL's ``learning by interacting'' during training and RL's responses to the environment based on rewards during testing. However, similar to supervised learning, RL also separates its training time from testing. Training an RL agent is similar to training a baby from birth to becoming an expert; only after the training completes, the grown agent is then ready to be tested in a new world. Training an RL agent may require numerous episodes, parallel to playing the same game many times, to understand the secret of winning. 

{\bf Bandits algorithms} study the problem of selecting actions from multiple choices each time to maximize a long-term reward. 
They might be the simplest form of RL, whose Markov Decision Processes (MDPs) are stateless. We can think of them as approximating the long-term return of taking an action with error correction and regularization. With the simplification in other components, they amplify the concept of exploration-exploitation trade-off in RL.   

Bandits algorithms are suitable for gain cumulative IS tasks. We can fund them in recommender systems, online advertising, multi-turn QA, DS, and eDiscovery. Here we mainly survey their use in the DS systems. For instance, \citet{ahukorala2015balancing} investigated how the rate of exploration in a bandits algorithm, LinRel~\cite{DBLP:journals/jmlr/Auer02}, would impact retrieval performance and user satisfaction.

The upper confidence bound (UCB) algorithm is a widely used and easy-to-implement method for exploration. It takes into account both the maximality of action and the uncertainty in the estimation. The action selection is made by 
\begin{equation}\label{eq:ucb}
    a_t = \argmax_a \left[ Q_t(a) + c \sqrt{\frac{\ln t}{N_t(a)}} \right]\,,
\end{equation}
where $Q_t(a)$ is the current estimation of the action value and $N_t(a)$ is the number of times action $a$ has previously been selected. The first term in Eq. \ref{eq:ucb} evaluates the optimality of the action, while the second term accounts for the uncertainty.
\citet{DBLP:conf/cikm/LiRM16} extended the UCB algorithm \cite{agrawal1995sample} for DS  where multiple queries were available simultaneously. Each arm is a query, and the task is to select a query from the pool at each step to maximize the total rewards. Its lexical similarity with existing queries estimated the action value of a new query. 

\citet{DBLP:conf/ictir/YangY17} used contextual bandits for DS. They used LinUCB~\cite{DBLP:conf/www/LiCLS10} to choose among query reformulation options, including adding terms, removing terms,  re-weighting terms, and stopping the entire search process. At each step, LinUCB selects an action based on 
\begin{equation}
    a_t = \argmax_a \left( {x}^T_{t,a} \hat{{\theta}}_a + \alpha \sqrt{ {x}^T_{t,a} {A}^{-1}_a {x}_{t,a}} \right)\,,
\end{equation}
where ${A}_a = {D}_a^T {D}_a + {I}_d$
and $\hat{{\theta}}_a = ({D}_a^T {D}_a + {I}_d)^{-1} {D}^T_a {c}_a$. Here 
$d$ is the dimension of the feature space, ${x}_{t,a}$ is the feature vector,  ${\theta}$ is a coefficient vector, ${D}_a$ is the matrix of $m$ training examples, and ${c}_a$ is the vector that holds corresponding responses.

\citet{DBLP:conf/aaai/WangWW17} proposed a factorization-based bandits algorithm for interactive recommender systems. They proved a sub-linear upper regret bound with high probability. They used observable contextual features and user dependencies to improve the algorithm's convergence rate and tackled the cold-start problem in recommender systems.

\citet{DBLP:conf/cikm/GeyikDMS18} applied bandits algorithm in talent search and recommendation. Candidates are first segmented into clusters by latent Dirichlet allocation (LDA) (\citet{DBLP:conf/nips/BleiNJ01}). They then adopted UCB to select a cluster for recommending talents. 

Exploration has been a long-lasting research theme in IR. The difference is that conventional IR methods do not model it explicitly. Instead, they use well-reasoning heuristics. Diversity and novelty are two heuristic goals thoroughly discussed for ad hoc retrieval. To name a few pieces of work,  \citet{jiang2012duplicate} penalized duplicated search results when ranking documents. \citet{bron2010university} diversified retrieval results based on the maximal marginal relevance (MMR) \cite{DBLP:conf/sigir/CarbonellG98} and latent Dirichlet allocation (LDA)   \cite{DBLP:conf/nips/BleiNJ01}. \citet{DBLP:conf/sigir/RamanBC13} identified intrinsically diversified sessions with linguistic features and then re-ranked documents by greedily selecting those that maximize the intrinsic diversity. 
\section{Value-Based RL Approaches} 
\label{rl:value-based}

Value-based RL associates each state (and action) with a value. Values are the expected long-term rewards  that an agent will get if starting from state $s$ or (state,action) pair $(s,a)$ with a given policy $\pi$. At the policy update step, the best policy is the one that takes the highest-valued state or (state, action) pair:
\begin{equation}
    \pi(s) \leftarrow \arg \max_a Q(s,a)\,.
\end{equation}
Therefore, value-based RL aims to arrive at high-valued states. The values can come from a single super high-valued state or many reasonably high-valued states. Consequently, it is essential to know (or approximate) the values of the states by interacting with the environment.

 We represent a value by {\it value function} $V$ for states, and by {\it Q-function} $Q$ for a (state, action) pair. We define $V(s)$ and $Q(s, a)$ similarly:
 \begin{equation}\label{eq:v_value}
 \begin{split}
     V(s) &= \mathbb{E}_{\pi}[G_t|s_t=s]\\
     &=\mathbb{E}_{\pi} \left[\sum_{k=0}^{\infty}\gamma^k r_{t+k+1} | s_t=s  \right]\,,
 \end{split}
\end{equation}
\begin{equation}\label{eq:q_value}
\begin{split}
     Q(s,a) &= \mathbb{E}_{\pi}[G_t|s_t=s, a_t=a]\\
     &= \mathbb{E}_{\pi} \left[\sum_{k=0}^{\infty}\gamma^k r_{t+k+1} | s_t=s, a_t=a  \right]\,.
\end{split}
\end{equation}
The two have a close relationship:
\begin{equation}\label{eq:v_and_q}
    V(s) = \sum_a \pi(a | s) Q(s, a)\,,
\end{equation}
\begin{equation}\label{eq:q_and_v}
    Q(s,\pi(s)) =  V(s)\,.
\end{equation}
The surplus between an action's Q-function value (often called ``Q-value'') $Q(s,a)$ and its state value $V(s)$ is also known as the {\it Advantage function A(s,a)}. We can think the advantage as the surplus value brought by an action over the average value brought by any action in the same state. We can obtain $V$, $Q$, and $A$  similarly.  Conventionally, we can use Bellman equation to calculate them. For instance, 
\begin{equation}
    V(s) = \sum_a \pi(a|s)\sum_{s',r}p(s',r|s,a)( r + \gamma V(s'))\,,
\end{equation}
where the quantity $r+\gamma V(s')$ is the value observed one step ahead. Early RL works \cite{DBLP:books/lib/SuttonB98} such as Value Iteration, Policy Iteration, and Temporal-Difference (TD) methods, including Sarsa and  Q-learning, are from this idea.  More recent approaches include the following. 

{\bf Monte Carlo Methods} sample a batch of trajectories, each from the beginning to the end of an episode, and then use these samples to calculate the average sample return for each state (and/or action). These sample returns are approximations to the value functions:   
\begin{equation} \label{eq:mc}
    Q(s, a) \leftarrow \frac{1}{N} \sum_{i=1}^N G^i_t [s_t=s, a_t=a]\,,
\end{equation}
where $G^i_t [s_t=s, a_t=a]$ is the return of the $t^{th}$ step in the $i^{th}$ sampled trajectory, and the agent receives state $s$ and takes action $a$. \citet{DBLP:conf/www/JinSW13} proposed a Monte Carlo method for a special kind of DS, multi-page search results ranking. When a user clicks the ``next'' button in a Search Engine Results Page (SERP) to view the remaining results, a DS algorithm detects this special user action and re-ranks the results on the next page before the user sees them. The method \citet{DBLP:conf/www/JinSW13} used a combination of document similarity and document relevance as states. The method estimated the state value functions via Monte Carlo sampling. The Discounted Cumulative Gain (DCG) score (\citet{DBLP:journals/tois/JarvelinK02})  was the reward. 
The entire ranked list of documents was the action. The action selection method greedily selected documents one by one to approximate an optimal ranking list. A simulated user conducted an experimental evaluation to examine the top $N$ returning documents and clicked on those of interest. The proposed algorithm was shown to outperform BM25 (\citet{DBLP:journals/ftir/RobertsonZ09}), maximal marginal relevance (MMR) (\citet{DBLP:conf/sigir/CarbonellG98}), and Rocchio (\citet{rocchio1971relevance}) on several TREC datasets.    


{\bf Temporal-difference (TD) learning} is another common family of value-based RL. TD learning also relies on sampling, but it does not sample the entire episode. Instead, it only samples one step ahead. TD learning calculates the difference between (a) its estimate to the current action-value, $Q(s, a)$, and (b) another estimate to the same thing but done in a different way -- $Q(s,a) = r + \gamma Q(s', a')$ --  which is estimated after looking one step ahead. The second estimate (b) is the ``target'', which is assumed to be better than (a) because it has observed the actual reward $r$ generated from the current state-action. The difference between the two estimates is known as the {\it TD error}. TD error works in RL similarly to dopamine works in animal learning. They are the trigger for the learning agent to begin adjusting the behavior. The best-known TD learning method is Q-learning. Q-learning is an off-policy method. It updates the action-value $Q(s,a)$ by: 
\begin{equation}\label{eq:td}
    Q(s, a) \leftarrow Q(s, a) + \alpha ( r + \gamma \max Q(s', a') - Q(s, a))\,,
\end{equation}
where $\alpha$ is the learning rate, $s, a$ are the current state and action, and $s', a'$ are the next state and action. The quantity  $r + \gamma \max Q(s', a')$ is the target,  $Q(s, a)$ is the current estimate, and $r + \gamma \max Q(s', a') -Q(s,a)$ is the TD error. 

Note that $Q(s', a')$ and $Q(s, a)$ are instantiations of the same function $Q$, with different inputs. We therefore can approximate the same function $Q$ for both of them with {\bf supervised machine learning}. Deep Q-Network (DQN) (\citet{DBLP:journals/nature/MnihKSRVBGRFOPB15}) is an example of this kind. DQN approximates the Q-function by minimizing the loss between the target Q-function and the estimated Q-function:
\begin{equation}
   \label{eq-dqn}
    L(\theta_i) = \mathbb{E}_{(s,a,r,s')} \left[ \left(r + \gamma \max_{a'} Q(s',a'| \theta_i^-)  - Q(s, a| \theta_i )\right)^2 \right]\,,
\end{equation}
where $r + \gamma \max_{a'} Q(s',a'| \theta_i^-)$ is the target Q-function, parameterized by $\theta_i^-$,  and $Q(s, a| \theta_i )$ is the predicted Q-function, parameterized by $\theta_i$.  Note that $\theta_i^-$ is kept a few steps behind $\theta_i$ to avoid the ``moving target'' problem (\citet{DBLP:journals/nature/MnihKSRVBGRFOPB15}). At each step, with probability $1-\epsilon$, DQN chooses the action with the highest action-value and equally likely the remaining actions. It then samples a batch of state transitions from a replay buffer to learn the $Q$-function again. DQN has successful applications in games, robotics, and many AI applications. 

Using value-based RL for DS can be traced back to the 2000s. \citet{DBLP:conf/cikm/Leuski00} defined the state as the inter-document similarities and action as the following document to examine. The $Q$ function,  parameterized by $\theta$, was updated via TD learning and the update rule was  
\begin{equation} \label{eq:q}
   \Delta \theta_t = \eta (r_{t+1} + \gamma Q_{\theta}(s_{t+1}, a_{t+1}) - Q_{\theta}(s_t, a_t)) \triangledown Q_{\theta}(s_t, a_t)\,,
\end{equation}
where $\eta$ is the learning rate.  \citet{DBLP:conf/trec/TangY17} adopted DQN to re-formulate queries at every time step and then input the query to an off-the-shelf retrieval function to get the documents. The state was as a tuple of current query, relevant passages, and index of the current search iterations in their work. The reward was the document relevance discounted by ranking and novelty. The actions were query reformulation options, including adding terms, removing terms, and re-weighting terms. \citet{DBLP:conf/www/ZhengZZXY0L18} extended DQN for news recommendation. They incorporated user behavioral patterns into the reward function. The state was the embedding vector for users, and action was the feature representation of the news articles. 
Both online and offline experiments have shown that this approach improved state-of-the-art recommender systems.

\section{Policy-Based RL Approaches} 
\label{sec:policy-based}

Policy-based RL is also known as gradient-based RL. When improving the policy, these methods update the policy based on gradient ascent --  updating the policy with a small step in the direction that the expected return would increase the fastest. This family of methods is from the policy gradient theorem \cite{DBLP:books/lib/SuttonB98}. The theorem says that the gradient of the expected return would go in the same direction as the policy's gradient; therefore, finding the optimal return would be equivalent to finding the optimal policy. 

Policy-based RL approaches directly learn policy $\pi$  by observing  rewards $r$, skipping the step of finding out state-values or action-values.  This class of RL assumes  policy $\pi$ can be parameterized, and the goal is to learn the best parameter $\theta$ that optimizes some performance measure $J$: 
\begin{equation}\label{eq:reinforce}
    J(\pi_{\theta})  = \mathbb{E}_{\tau \sim \pi_{\theta}(\tau)} \left[ r(\tau) \right] = \int \pi_\theta (\tau) r(\tau) d\tau \,,
\end{equation}
where $\tau$ is a trajectory from time steps 1 to $T$, i.e. $s_0,a_0,r_1,s_1,a_1,r_2,...,s_{T-1},a_{T-1},r_T,s_T$. 


The most classic policy-based RL algorithm is REINFORCE (\citet{williams1992simple}). It is still widely used by AI applications, including dialogue systems and robotics. REINFORCE is a Monte Carlo policy gradient method. It first generates sample trajectories, then computes the return for each time step and updates the policy parameter $\theta$. Using a gradient ascent process, it updates its policy each time and arrives at $\theta$ that maximizes $J(\pi(\theta))$: 

\begin{equation}
\begin{split}
    \nabla_\theta J(\theta) 
    &= \int \nabla_\theta \pi_{\theta}(\tau) r(\tau) d\tau \\
    & \approx \frac{1}{N} \sum_{i=1}^N \left(\sum_{t=1}^T \nabla_\theta \log \pi_\theta (a_{i,t} | s_{i,t}) \right) \left( \sum_{t=1}^T r(s_{i,t}, a_{i,t}) \right)\,, \\ 
\end{split}
\end{equation}
\begin{equation}
    \theta \leftarrow \theta + \alpha \nabla_\theta J(\theta)\,,
\end{equation}
where $\alpha$ is the step size each time the gradient changes, and the method updates the gradient in the direction of the return's gradient. The algorithm \ref{algo:reinforce}details the REINFORCE algorithm. 

Other example policy-based RL algorithms include Trust-Region Policy Optimization (TRPO)~\cite{DBLP:conf/icml/SchulmanLAJM15} and Proximal Policy Optimization (PPO)~\cite{DBLP:journals/corr/SchulmanWDRK17}. They are state-of-the-art RL algorithms, and many applications use PPO for its simple implementation. Besides, 
We can combine value-based and policy-based methods. When they are combined, the new algorithms are called  {\it actor-critic}. Actor-critic ways are highly effective. Examples include A3C \cite{DBLP:conf/icml/MnihBMGLHSK16} and DDPG \cite{DBLP:journals/corr/LillicrapHPHETS15}.

\begin{algorithm}[t]
\SetAlgoLined
 initialization\;
 \While{True}{
  Generate a trajectory $s_0,a_0,r_1,s_1,a_1,...,s_{T-1},a_{T-1},r_T$ following $\pi_\theta$\;
  \For{each step $t=0,1,2,...,T-1$}{
  $G_t =\sum_{k=0}^{T-t-1}\gamma^k r_{t+k+1}$\;
  $\theta \leftarrow \theta + \alpha \nabla_\theta J(\theta)$\; 
  }
 }
 \caption{REINFORCE by \citet{williams1992simple}.}\label{algo:reinforce}
\end{algorithm}


In 2015, \citet{DBLP:conf/ictir/LuoDY15}  proposed direct policy learning for DS. In their work, each search iteration consists of three phases--browsing, querying, and ranking. Its parameters parameterize each phase. The action is sampled only in the ranking phase from search results generated by an ad hoc retrieval function. But previous phases are passed in as observations for the final action, which are updated based on the expected return's gradient. The algorithm \ref{algo:pg_session}shows the direct policy learning algorithm for DS.   



\begin{algorithm}[t]
$\theta \leftarrow random(0,1)$ \;
\Repeat{$\Delta \theta < \epsilon$ or history set is empty}{
Sample history $h$ from history set\;
$q_0, D_0, C_0, T_0 \leftarrow \emptyset$ where $q_*$ is the query, $D_*$ is the set of retrieved documents. $C_*$ are the clicks, $T_*$ is the dwelling time \;

\For{$t$ in range($len(h)$)}{
$o_{rank} \leftarrow D_{t-1}, n_1 \leftarrow o_{rank}$\;
The user takes action $a_{browse}$\;
$o_{browse} \leftarrow (C_t, T_t), n_2 \leftarrow o_{browse}$\;
$r(t, h) \leftarrow CalculateReward(D_t', o_{browse}, h)$\;
The user takes action $a_{query}$\;
$o_{query} \leftarrow GetQueryChange(q_t, q_{t-1})$\;
$n_3 \leftarrow (o_{rank}, o_{browse}, o_{query})$\;

Sample a search engine action $a_{rank} \approx P(a_{rank} | n_3, \theta) $\;
$D_t' \leftarrow DocRanking(a_{rank})$\;
$\Delta \theta \leftarrow UpdateGradient(r(t, h), D_t', n_3, \theta)$\;
$\theta \leftarrow \theta + \Delta \theta$\;
}

}
\caption{Direct Policy Learning for Dynamic Search, by \citet{DBLP:conf/ictir/LuoDY15}.}\label{algo:pg_session}
\end{algorithm}

Policy-based RL methods are also popular in dialogue systems, multi-turn question answering (QA), and query reformulation.  \citet{DBLP:conf/acl/DhingraLLGCAD17} proposed KB-InfoBot, a multi-turn dialogue agent for knowledge base-supported movie recommendation. They tracked user intents in a belief tracker in their work, which indicated slot probabilities in utterances. They also used a soft lookup function modeling a posterior distribution of user preferences for all the items. The belief tracker output and the lookup results together constituted the RL agent's state. The action was to ask the user clarification questions about the target item or inform the final answer. They optimized the policy by maximizing the returns. 

In 2019, \citet{das2018multistep} proposed a multi-hop reasoning method for open-domain QA. It had two agents, a retriever, and a reader. The two agents interacted with each other. The retriever's job was to find relevant passages from which the reader extracts answers to the question. The query was a feature vector, which was initialized by embedding the input question. A multi-step reasoner reformulated the query -- i.e., modified the query vector -- based on the reader's output and the question vector. The state was the input question and the inner state of the reader. The action was to select which paragraphs to return. The reward was measured by how well the reader's output matched with ground-truth questions. We can train the model with REINFORCE; the algorithm is in Algorithm \ref{algo:multi_qa}.

\citet{DBLP:conf/emnlp/LiMRJGG16} applies RL to generate dialogues between two virtual agents. The state was their previous round of conversation, $[p_i, q_i]$. Action was the next utterance with arbitrary length $p_{i+1}$. They measured the reward in terms of the utterance's informativity, coherence, and ease with which We can answer it. They parameterized the policy using a Long Short-Term Memory (LSTM)  encoder-decoder neural network. The system also used REINFORCE (\citet{williams1992simple}).

\begin{algorithm}[t]
\SetAlgoLined
${q_0} \leftarrow encode\_query(Q)$ where $Q$ is the input question\;
\For{ $t$ in range($T$)}{
    ${p_1, p_2, ..., p_k} \leftarrow retriever.score\_passages({q_t})$ \;
    $P^{1...k}_{text} \leftarrow reader.read({p_1, ... , p_k})$\;
    $a_t$, ${score}_t$, ${S_t}  \leftarrow reader.read(P^{1...k}_{text}, Q)$, where $a_t$ is the answer span, ${score}_t$ is the score of the span, ${S_t}$ is the reader state\;
    $\bm{q_t} \leftarrow reasoner.reason({q_t}, {S_t})$\;
    }
    \Return answer span $a$ with the highest score
\caption{Multi-step reasoning for open-domain QA, by \citet{das2018multistep}}\label{algo:multi_qa}
\end{algorithm}

\citet{DBLP:conf/emnlp/NogueiraC17} proposed RL-based query reformulation. Its state was the set of retrieved documents,  and the action was to select terms to formulate a query. The original query and candidate terms, along with their context, are fed into a siamese neural network, whose output is the probability of adding a term into the new query. They also trained this system with REINFORCE. \citet{DBLP:conf/wsdm/ChenBCJBC19} scaled REINFORCE to vast action space and proposed Top-K off-policy correction for recommender systems. \citet{10.1145/3366423.3380294} also used REINFORCE for DS with the AOL query log.  

In 2017, \citet{DBLP:conf/emnlp/AissaSD18} proposed a conversational search system by translating information need into keyword queries. They trained a Seq2Seq network with REINFORCE. The Seq2Seq network acted as the policy. Its input was the information need, expressed as a sequence of keywords, and the output was a sequence of binary variables indicating whether to keep a term in the query. The reward was the mean average precision (MAP) of the search results obtained by the new query. The task was to generate topic titles from some descriptions, using the TREC Robust Track (\citet{DBLP:conf/trec/Voorhees04b}) and Web Track (\citet{DBLP:conf/trec/Hawking00}) datasets. Compared with supervised learning approaches, training the Seq2Seq model with RL achieved the best performance.

\section{Model-Based RL Approaches}

In RL, the word  ``model''  specifically refers to the state transition function $M$ and sometimes includes the reward function $R$. It represents the dynamics of the environment: 
\begin{equation}
    s', r' \leftarrow Model(s,a)\,,
\end{equation}
where $a$ is the action taken by the agent after it perceives state $s$; the model outputs the resultant state as $s'$ and the immediate reward as $r'$. 

\begin{figure}[t]
    \centering
    \includegraphics[width=0.3\linewidth,height=0.22\linewidth]{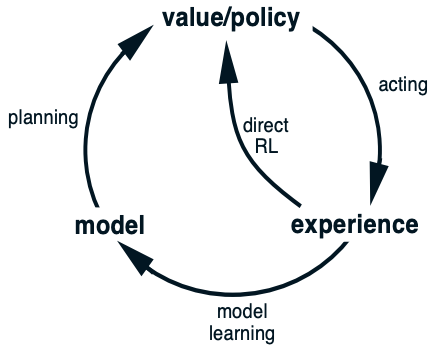}
    \caption{Model-based RL. ``Model'' in reinforcement learning has a special meaning to only refer to the model that represents state transitions (and sometimes include a reward model too). It can have two parallel lines of learning. First,  it is the model-free RL (also known as direct RL) that we presented earlier. For instance, policy-based RL and value-based RL. This line of RL learns the policy directly from getting environment's rewards (feedback) after taking an action, i.e., learns from experience.  Another line of learning requires learning the state transition model (the ``model'' in RL) from the experiences, and then make use of this learned transition model to plan the next action, or simulating rewards (if the reward model is also learned) to augment the experiences. Here the models are mostly learned by supervised learning methods. The model can be thought of an estimation to the real environment, which can move on to the next state and gives rewards to the RL agent. The second line of learning is called indirect RL, since the agent does not directly receive rewards from the environment, but from the learned model. The two lines of learning can happen at the same time or in sequence; they can be complementary to each other. Graph from \citet{DBLP:books/lib/SuttonB98}. }
    \label{fig:model2}
\end{figure}

Both value-based and policy-based methods do not utilize the state transition model $M$. Thus they are called {\it model-free} methods. When an RL method knows or learns the model $M$, and then makes plans with $M$ when taking actions, we call them {\it model-based} RL methods. Figure \ref{fig:model2} shows the model-based RL paradigm.

The transition model $M$ can be learned from data. The learning can happen (1) before, (2) alternating with, and (3) together with policy updates. First, learn the transition model before policy updates appear in early work. They include optimal control and planning.  Monte-Carlo Tree Search (MCTS) is an example of planning. AlphaGo used it in combination with actor-critic \cite{DBLP:journals/nature/SilverHMGSDSAPL16} and beat human Go champions in 2016.   
Second, the transition model $M$ can also be learned alternately with the policy. Dyna \cite{DBLP:books/lib/SuttonB98} is a classic algorithm of this kind. 
Third, It can learn the transition model by calculating a gradient through both the model and policy proposed in PILCO \cite{DBLP:conf/icml/DeisenrothR11} and Guided Policy Search (GPS) \cite{DBLP:conf/icml/LevineK13}.

Value Iteration (VI) is perhaps the simplest form of model-based RL. We can derive the method from the Bellman equation. It approximates the optimal state-value as
\begin{equation}\label{eq:value_iteration}
    V(s) = \max_a R(s, a) + \gamma \sum_{s'} M(s'| s, a) V(s')\,,
\end{equation}
where $s$ is the current state, $s'$ is the next state, and $M$ is the transition function and can be represented by a tabulate. In 2013, \citet{DBLP:conf/sigir/GuanZY13} proposed the Query Change Model (QCM) for DS. It is a term-weighting method optimized via value iteration. QCM defined states as queries and actions as term-weighting operations, such as increasing, decreasing, and maintaining term weights. The level of weight to be adjusted as determined by syntactic changes in two adjacent queries. QCM denoted the state transition function as a query-change probability, $P(q_i, q_{i-1}, D_{i-1}, a)$. Here $q_i$ is the $i^{th}$ query, $D_i$ is the set of  documents  retrieved by $q_i$, and $a$ is the query change action. Different query change actions would result in different $q_i$.  
To calculate the state transition, the two adjacent queries $q_i$ and $q_{i-1}$ were first broken into tokens. The probability of $P(q_i, q_{i-1}, D_{i-1}, a)$ was then calculated based on how each of the tokens would appear in $q_i$, $q_{i-1}$ and $D_{i-1}$. QCM then estimated the state-value for each document and used it to score the document: 
\begin{equation}\label{eq:qcm}
\begin{split}
     Score(q_i, d) &= P(q_i, d) + \\
     & \gamma \sum_a P(q_i, q_{i-1}, D_{i-1}, a) \max_{D_{i-1}} P(q_{i-1} | D_{i-1})\,,
\end{split}
\end{equation}
where the first term, $P(q_i, d)$, measures the relevance between document $d$ and the current query $q_i$, with a ranking score provided by a standard ad hoc retrieval function. The second term is complex: $P(q_i, q_{i-1}, D_{i-1}, a)$ is the transition probability from $q_{i-1}$ to $q_i$, given the query change $a$; $\max_{D_{i-1}} P(q_{i-1} | D_{i-1})$ is the maximum possible state value generated from the retrieval that has just happened.   

\begin{figure}[t]
    \centering
    \includegraphics[width=0.55\linewidth]{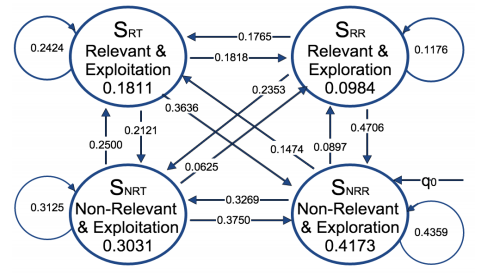}
    \caption{State transitions in Win-Win search. The four states are jointly designed along two orthogonal aspects in an information seeking process. The aspects are relevance and desire to explore. Note these aspects and states are decision-making states from the user's perspective. The graphic shows the state probabilities and the state transition probabilities learned from past search logs. Picture is adapted from \cite{DBLP:conf/sigir/LuoZY14}.}
    \label{fig:win_win_transition}
\end{figure}

In 2014, \citet{DBLP:conf/sigir/LuoZY14} proposed the win-win search method to study DS as Partially Observable MDPs (POMDPs). Two agents, the search engine agent and the user agent participated in the process. The state was modeled as a cross-product of (a) whether it was relevant and (b) whether it was exploring. Actions on the system side included changing term weights and opting in certain retrieval functionalities. Actions on the user side included adding or removing query terms. The model, aka the state transition probabilities, was estimated by analyzing query logs. For instance, how likely a user's decision-making state would go from relevance \& exploration ($S_{RR}$) to relevance \& exploitation ($S_{RT}$). They used the estimated model to select actions that optimize the expected return. They also manually annotated the state of queries and used them to train the model. Examples of the state transition probabilities are shown in Figure \ref{fig:win_win_transition}, where the likelihood of moving from state $S_{RR}$ to state $S_{RT}$ is $0.1765$.

In 2016, \citet{DBLP:conf/sigir/ZhangZ16} proposed to use  model-based RL for search engine UI improvement. It also used value iteration. The task was to choose an interface card at each time step and show it to the user. Its state was the set of interface cards that They had not yet presented. The action was to choose the next card to show. That is, $s_t = \{e: a_{t'} \neq e, \textrm{ } \forall t' < t  \}$, where $e$ is an interface card. 
The transition model was to exclude an element from the set:
\begin{equation}
    s_{t+1} = M(s_t, a_t) = s_t \setminus \{ a_t\}\,.
\end{equation}
The Markov Decision Processes (MDPs) were solved using value iteration (Eq. \ref{eq:value_iteration}). Simulation and a user study have shown that the method could automatically adjust to various screen sizes and satisfy users' stopping tendencies. 

In 2018, \citet{DBLP:conf/sigir/FengXLGZC18} proposed an actor-critic framework for search result diversification. Its state was a tuple $[q, Z_t, X_t]$, where $q$ was the query, $Z_t$ the current ranked list, and $X_t$ the set of candidate documents. The action was to choose the next document. The transition model was about to append a document at the end of $Z_t$ and remove it from $X_t$: 
\begin{equation}
\begin{split}
    s_{t+1} &= [q, Z_{t+1}, X_{t+1}] = T(s_t, a_t) \\ 
    &= T([q, Z_t, X_t], a_t) 
    = [q, Z_t \oplus a_t , X_{t} \setminus \{a_t\}]\,,
\end{split}
\end{equation}
where $\oplus$ is the appending operator, they then used  Monte-Carlo Tree Search (MCTS) by \citet{DBLP:conf/aiide/ChaslotBSS08} to evaluate the model and suggest the policy. 

Because model-based RL separates model learning and policy learning, it then provides opportunities to inject domain-specific knowledge and human expertise into the model and then helps with the policy learning. Much RL research is ongoing along this direction.

%


\section{Imitation Learning}
\label{imitation}

The process of IS requires a search engine to take a sequence of actions. These actions form a search trajectory (or search path). Learning from how human experts make these decisions is a natural choice to decide the best actions. The agent learns from sequences of human expert-generated actions (with labels) and minimizes the difference between the ML model and the labeled data. The training data is the recorded history of past action sequences by human experts. 

These supervised methods for sequential decision-making are called {\it imitation learning} (IL)~\cite{DBLP:journals/ftrob/OsaPNBA018}, also known as {\it learning by demonstration} or {\it programming by examples}. Applications such as self-driving cars heavily use imitation learning. In the well-known DAgger algorithm~\cite{DBLP:journals/corr/abs-1011-0686}, the algorithm collects state-action pairs $(s, a)$ as training data from a human expert demonstration. The algorithm then fits a function $f(s)=a$, which would be the policy for the agent. We can learn the policy properly by minimizing the error between the estimated actions and the expert actions. 

In IS, the human demonstration is the query logs, and imitation learning teaches a sequential decision-making process from query logs~\cite{DBLP:conf/cikm/YangGC18,DBLP:conf/www/RenNMK18,DBLP:conf/nips/VaswaniSPUJGKP17,DBLP:conf/sigir/Mitra15}. A query log usually contains past queries, retrieved documents, time spent examining the documents, and user clicks. Query logs are largely available within commercial search engines but rarely shared for academic research due to privacy concerns. Academic search logs can be found in the TREC Session Tracks (2010-2014)  (\citet{DBLP:conf/sigir/CarteretteCHKS16}) and the TREC Conversational Assistant (CaST) Track (\citet{dalton2020trec}).
The TREC Session Track invited academic participating systems \cite{xue2014ictnet,huurnink2012university} to learn from past queries in the same session to improve the search for the current query. The TREC CaST Track provided search logs developed by third-part expert users. The search log includes the history of queries and gold standard relevant document/answer at each query, establishing an expert demonstration for the learning agent. Search systems can learn from this demonstration and retrieve relevant documents/passages that are responsive to each query. 

The sequence-to-Sequence (Seq2Seq) model is so far the most used imitation learning method for IS. Seq2Seq is a recurrent neural network model that encodes a sequence in one domain and decodes the sequence in another. It can predict the sequence of clicks, query reformulations, and query auto-completion (QAC). For instance, \citet{DBLP:conf/sigir/BorisovWMR18} predicted clicks on the Yandex Relevance Prediction dataset\footnote{\url{https://academy.yandex.ru/events/data_analysis/relpred2011/}} with a Seq2Seq deep neural network. Its input sequence is the sequence of queries and retrieved items; the output sequence is a sequence of binary markers for the items that are clicked or not. The human demonstration is to improve the Seq2seq model. \citet{DBLP:conf/www/RenNMK18} also employed a Seq2Seq model to generate new queries given past queries and search results. Its model used two layers of attention networks \cite{DBLP:journals/corr/BahdanauCB14}. The state consisted of previous queries and  user behavior data in past interactions. \citet{DBLP:journals/corr/SordoniBVLSN15} used a hierarchical recurrent encoder-decoder for query suggestion, which performed the encoding at both the individual word-level and the session-level.  \citet{DBLP:conf/cikm/DehghaniRAF17} proposed a customized Seq2Seq model for query auto-completion and query suggestion. They added a copying mechanism in the decoder to directly select query terms from the query log instead of generating new query terms.   

Researchers also explored other supervised methods for imitation learning. 
For instance, estimating document relevance with logistic regression by \citet{huurnink2012university} and  with SVMRank by \citet{xue2014ictnet}. In \citet{DBLP:conf/sigir/Mitra15}, they  used a convolutional latent semantic model \cite{DBLP:conf/www/ShenHGDM14}
 to study a distributed query representation for query auto-completion. For example, $v(\mbox{new york}) + v(\mbox{newspaper}) \approx v(\mbox{new york times})$, where $v(*)$ is the word embedding vector.
\citet{DBLP:conf/cikm/YangGC18} used a deep self-attention network \cite{DBLP:conf/nips/VaswaniSPUJGKP17} for query reformulation. They modeled a two-round interaction between user and search engine and formulated new queries based on words in past queries.

\section{Open-Ended RL}
\label{open-ended} 

Open-ended RL is ongoing RL research~\cite{poet,Pfeiffer,Standish01open-endedartificial,lastchallenge}. The term ``open-ended'' here suggests the task domain (environment) is unbounded. The environment will no longer be a closed world. However, it will adapt and evolve to other forms as the task develops. The current study of open-ended RL focuses on domain randomization~\cite{DBLP:conf/sigir/ChenTY20} and evolutionary methods~\cite{evolution}. 
 
\citet{DBLP:conf/sigir/ChenTY20} used domain randomization in DS. They automatically generated simulated environments while training RL agents. They created newly generated environments from the original task environments. The RL agents' training used both original and newly-generated environments. It increased the diversity of the training and prepared the agents better and more robust. 

In the context of IS, open-ended RL can help with exploratory search where the task domain can change significantly as the user explores. It can also help with the co-evolvement of both the user and the search system~\cite{evolution}. However, to the best best of our knowledge, there has not been much research in IR/IS with open-ended RL as solutions. 
\section{Evaluation} 
\label{eval}

We have reviewed IS tasks and methods and re-classified them. The methods need evaluation for their effectiveness. Although evaluating a dynamic task is challenging, 
TREC and the Conference and Labs of the Evaluation Forum (CLEF) have devoted much effort to the evaluation testbeds for IS. These efforts include the TREC Interactive Tracks from 1997 to 2002 \cite{DBLP:conf/trec/Voorhees02,trecinteractive1998,trecinteractive2005}, TREC Session Tracks from 2011 to 2014 \cite{TREC:2010:Session,TREC:2011:Session,TREC:2012:Session,TREC:2013:Session,TREC:2014:Session,DBLP:conf/sigir/CarteretteCHKS16}, TREC Dynamic Domain Tracks from 2015 to 2017 \cite{DBLP:conf/trec/YangTS17,DBLP:conf/trec/YangS16,DBLP:conf/trec/YangFS15}, CLEF Dynamic Search Lab in 2018 \cite{DBLP:conf/clef/KanoulasAY18} and the recent TREC Conversational Assistant (CAsT) Track \cite{dalton2020trec} starting from 2019. 

{\bf The Campaigns.} The TREC Interactive Track's setting \cite{DBLP:conf/trec/Voorhees02,trecinteractive1998,trecinteractive2005} is perhaps the closest to information seeking scenarios in real life. Participating teams carried out the search tasks in their home institutions. Human users and search engines were both parts of a team. In the end, each participating team submitted the final search results for evaluation. This Track allowed live interactions between the user and the search engines. However, the assessment did not separate the work from the user and the result from the search engines. It is thus difficult to tell how effective the search engine is without manual influence. Also, no one could repeat the evaluation afterward, which prevents follow-up research. 

\begin{figure} [t]
    \centering
    \includegraphics[width=0.8\linewidth]{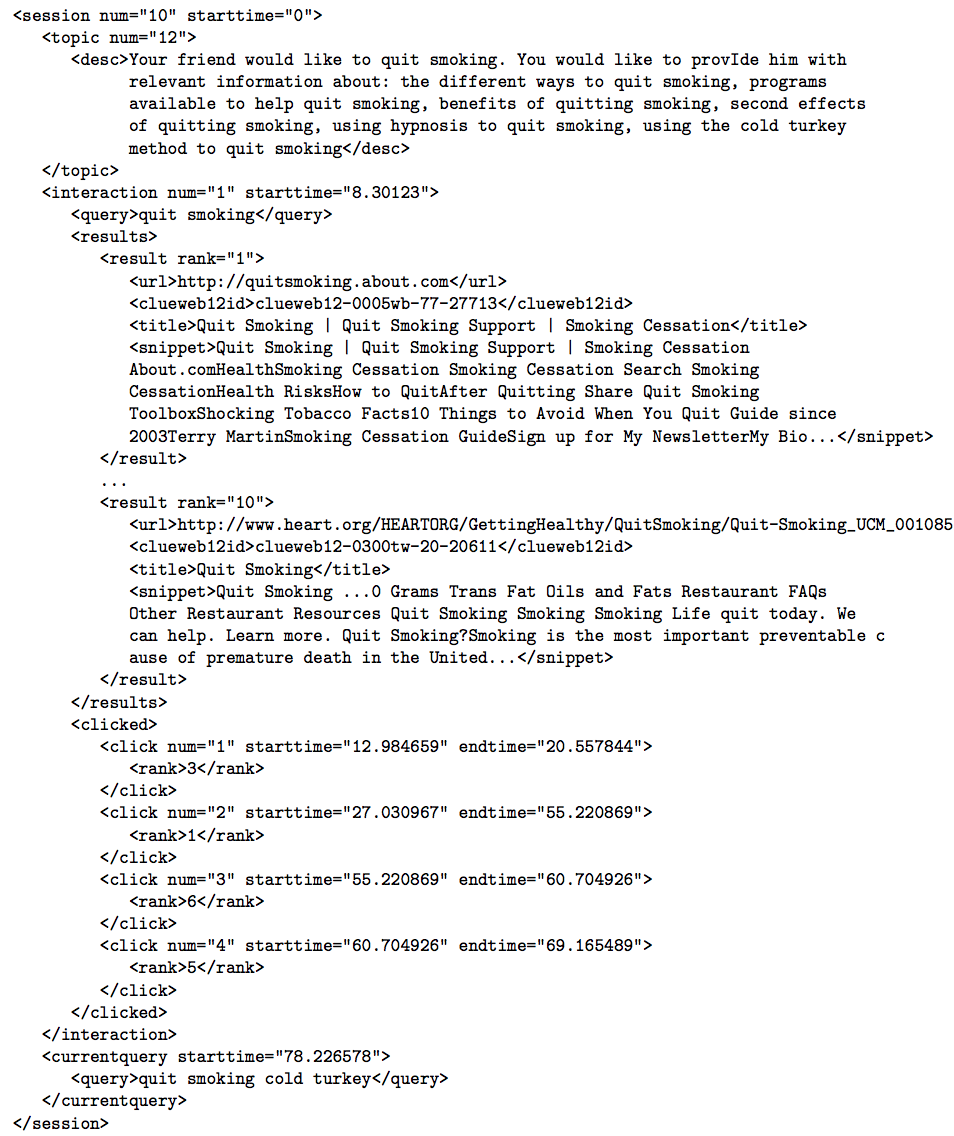}
    \caption{Example search log used in TREC Session Tracks. The TREC Session Tracks provided search logs as training data. Each session's log shows the main search topic (annotated with the ``<desc>'' tag), every query terms issued by a real human user (annotated with the ``query'' tag), content of the top ten returned documents by a search engine that the Track organizers used, clicked documents and their ranks, time that a click happens, and all clock time information of user actions. Multiple iterations for the same search topic are recorded and stored in one session. The last query is left open with no retrieval results, expecting the participating search systems to fill it up. Content from \citet{TREC:2014:Session}. }
    \label{fig:searchhistory}
\end{figure}

The TREC Session Track \cite{TREC:2010:Session,TREC:2011:Session,TREC:2012:Session,TREC:2013:Session,TREC:2014:Session,DBLP:conf/sigir/CarteretteCHKS16} is a log-based evaluation. Its setting is very similar to what is available in commercial search engine companies. The Track provided participating systems query logs containing search histories and expected to retrieve search results for a new (current) query. The search logs included recorded queries, retrieved URLs, clicks, and dwell time. Figure \ref{fig:searchhistory} shows an example search log used in the Track. Clicks and visit time can be extracted from the search logs and used as implicit feedback. There was no real-time interaction between the user (whose behaviors are historical) and the search engine (the participating systems). However, the evaluation is repeatable; thus, it had inspired follow-up research even when the evaluation period was over. 

The TREC Conversational Assistance (CAsT) Track \cite{dalton2020trec} is another log-based evaluation. Participating systems were provided sequences of <question, canonical answer document> pairs and expected to retrieve answer documents at each conversational turn. The series of conversations is fixed and recorded from past search histories. The feedback (or supervision) is in the form of the canonical answer at each turn. There was no real-time interaction between the user (whose questions are historical) and the search engine (the participating systems). However, the evaluation is repeatable, and we could collect more training data in the following years. 

\begin{figure} [t]
    \centering
    \includegraphics[width=0.95\linewidth]{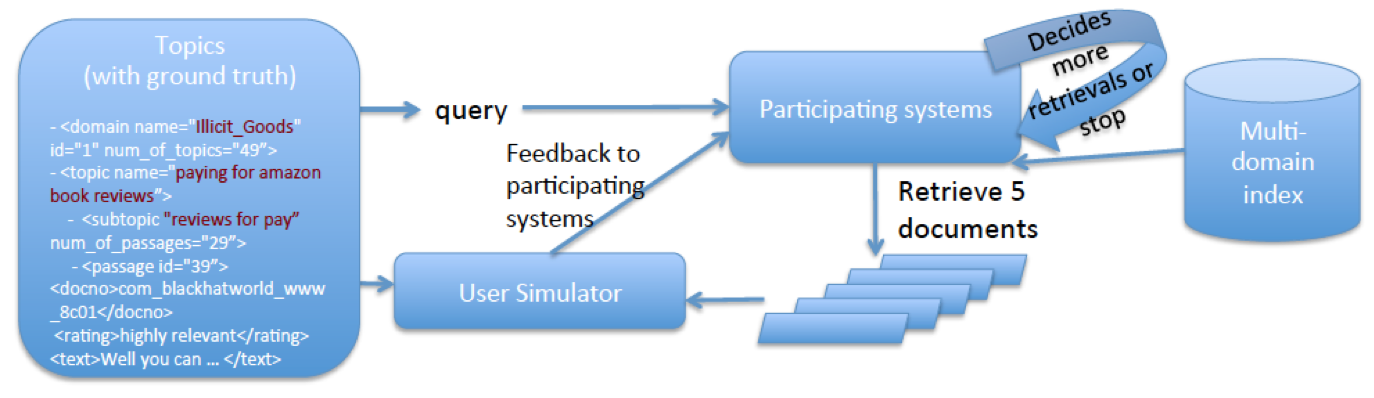}
    \caption{Task Setting in TREC Dynamic Domain Tracks. TREC DD supports simulation-based evaluation. It uses a simulated user, called ``jig'', who reads/processes the returned search documents (or passages) from a participating search system, and gives the search system its immediate feedback. The Feedback is a set of multiple graded ratings to each returned document, one rating for each relevant passage that the document has. The feedback also indicates to which subtopic the passage is relevant. The idea is that the simulated user acts like a human reader to tell the search engine, ``I like this paragraph in this document that you retrieved. It is relevant to this of my sub-information-need with a level of relevance of x (x is an integer from 1 to 4). The search system gets the feedback and goes on to find another set of documents, hoping make good use of the feedback to help the next run of retrieval and finish the entire search task quicker. Each time the search engine is only allow to retrieve a short list of five documents. This is to mimic the Web search scenario and assume a low cognitive load at the user side. The search system therefore needs to be smart enough to select a short list of documents and find all relevant passages to cover all subtopics use as few iterations as possible. The ground truth data and all relevance ratings was created before any search system interacting with the simulated user and stored in a database. They serve as the simulated user's knowledge base. This task supports repeatable experiments and open comparisons among search systems. The data and simulated user can be obtained from \url{https://github.com/trec-dd/trec-dd-jig}. Picture from \citet{DBLP:conf/trec/YangFS15}.}
    \label{fig:dd}
\end{figure}

The TREC Dynamic Domain (DD) Tracks \cite{DBLP:conf/trec/YangTS17,DBLP:conf/trec/YangS16,DBLP:conf/trec/YangFS15} setting is similar to the TREC Interactive Track. The only difference is it replaces the human user with a simulator. Its setting is, therefore, more like a game. TREC DD's simulator provides real-time feedback to a search engine. The input includes (1) relevance ratings to returned documents and (2) information highlighting which passages are relevant to which subtopics. The search engine can adjust its search algorithm and find relevant documents at each iteration. The canonical feedback from the simulator was from third-party annotators before the evaluation period. By using a simulator, TREC DD supports live interactions and reproducible experiments. However, its game-like setting may appear to be artificial. Figure \ref{fig:dd} illustrates TREC DD's setting.

The CLEF Dynamic Search Lab in 2018 \cite{DBLP:conf/clef/KanoulasAY18} is similar to and has further developed the TREC DD Tracks. It also used simulation to carry out the evaluation. Moreover, the assessment for two agents, a Q-agent, reformulates queries, and an A-agent retrieves documents. The Lab evaluated both query suggestions and search results composition. Because of using simulators, the Lab can support live interactions and reproducible experiments. 

{\bf Metrics} These evaluation campaigns have used a range of evaluation metrics over the years. The metrics are (1) ad hoc retrieval metrics, (2) session-based, (3) effort-based metrics, and  (4) indirect evaluation metrics. 
The first group of metrics concerns the effectiveness of the result from information seeking. They use the same metrics for one-shot ad hoc retrieval. For instance, TREC Interactive Tracks used aspect precision. They used aspect recall,  TREC Session used normalized discounted cumulative gain (nDCG), TREC DD used $\alpha$-DCG \cite{DBLP:conf/sigir/ClarkeKCVABM08} and nERR-IA\cite{DBLP:journals/ir/ChapelleJLVLW11} . Note that most of these metrics take into account the complex structure within the information seeking topics. Therefore metrics addressing subtopic structures (aspect precision, aspect recall, $\alpha$-DCG, nERR-IA) are widely used. 

The second group of metrics views information seeking as a process and measures the entire process. For instance, Cube Test (\citet{DBLP:conf/cikm/LuoWYH13}), session DCG (\citet{DBLP:conf/ecir/JarvelinPDN08}), and Expected Utility (\citet{DBLP:conf/ictir/YangL09}). Among them, the Cube Test \cite{DBLP:conf/cikm/LuoWYH13} evaluates the speed of effective retrieval of a DS system, i.e., the ratio between the gain and the time spent. Expected utility \cite{DBLP:conf/ictir/YangL09} measures the ``net gain'' of the search process, i.e., the amount of gain subtracting the user effort. Session DCG is a cumulative gain over the entire session. These metrics evaluate the performance of a dynamic search system in terms of the relevance of information acquired and the effort the user put into the search process, based on different user models~\cite{DBLP:conf/ictir/TangY17}.

The third group of metrics recognizes that information seeking is a process that takes effort. These metrics include  Time-Biased Gain (TBG) \cite{DBLP:conf/sigir/SmuckerC12} and U-measure \cite{DBLP:conf/sigir/SakaiD13}. They look into the user's efforts in an IS process by considering that a user may spend different time on different parts of the search results. 

The fourth group of evaluation is indirect evaluation. CLEF Dynamic Search Lab~\cite{DBLP:conf/clef/KanoulasAY18} evaluates reformulated queries indirectly by measuring how relevant those queries retrieve the documents. It does not compare reformulated queries with human-generated queries. 

\begin{table}[t]
    \centering \small
    \begin{tabular}{c|l}
    \toprule
        Topic (id: DD17-49)& Kangaroo Survival   \\
    \midrule
        Subtopic 1 (id: 464) & \makecell[l]{Road Danger \\ - - 7 Relevant Passages}\\
        Subtopic 2 (id: 462)& \makecell[l]{Effect on Land \\ - - 15 Relevant Passages}\\
        Subtopic 3 (id: 463)& \makecell[l]{Monetizing Kangaroos \\ - -  65 Relevant Passages}\\
        Subtopic 4 (id: 459) & \makecell[l]{Kangaroo Hunting  \\ - - 17 Relevant Passages}\\
        Subtopic 5 (id: 552) & \makecell[l]{Other Kangaroo Dangers \\ - - 4 Relevant Passages}\\
        Subtopic 6 (id: 460) & \makecell[l]{Protection and Rescues \\ - - 57 Relevant Passages}\\ 
    \bottomrule
    \end{tabular}
    \caption{Example Search Topic in TREC DD. Each search topic has a two-level topic hierarchy. At the top-level, the main search theme of the topic is displayed. In this example, it is ``kangaroo survival''. At the second-level, several subtopics are organized under the main theme. They are all about the same theme, but address different aspects of it. in this example, they include road danger, effect on land, monetizing kangaroos, kangaroo hunting, protection and rescues, and other kangaroo dangers.  All the main theme (the topic) and the subtopics are assigned a unique query identification number. For each subtopic, TREC DD has ground truth relevant documents and passages data annotated. These ground truth data can be easily aggregated for the entire search topic. How many relevant passages for each subtopic are noted, too. All groud truth documents and passages can be accessed from \url{http://infosense.cs.georgetown.edu/annotation/admin/admin.cgi}. }
    \label{tab:example_topic}
\end{table}

Evaluating a dynamic process is challenging. It is because the relevance is situational \cite{borlund2003concept} in the IS process~\cite{DBLP:conf/cikm/MedlarG18}. However, a large amount of effort spends by the IR community on understanding this dynamic process. We gain promising insights. For instance,~\citet{DBLP:conf/sigir/HuffmanH07} found that relevance judgment correlated well with user satisfaction at the session-level. ~\citet{DBLP:conf/sigir/LiuLMLM18} found that the last query in the process played a vital role.

{\bf Structures within a Search Topic} Most search topics in these evaluation campaigns have complex within-topic structures. Because if a topic is simple, ad hoc retrieval can then resolve it, and no need to enter the multi-step IS the process. 

The simplest topic structure is a two-level hierarchy. At the top level is the central search theme, and at the lower level are the subtopics. Table \ref{tab:example_topic} shows an example search topic with a two-level hierarchy. Its main information need is {\it to find factors that would affect kangaroo survival} and the six subtopics including {\it traffic accidents} and {\it illegal hunting}. 
 
A search path is the ordered visitation to the nodes in a topic hierarchy (or, more generally, a topic topology). Search paths can only be obtained dynamically in real-time interactions or simulations, while the topic structures are static. In both TREC Session and TREC CAsT, they logged and showed the participating systems a particular search path without revealing the underlying topic structure. In TREC DD and CLEF DS Lab, the participating systems can see the topic structures. However, they cannot explicitly use them in their algorithms. Each time, the real-time simulation could show a different search path to the search system. 


However, it is worth-noting that a task's internal structure can affect if the task would be ``task completion'' or ``cumulative gain.'' Suppose a task's internal structure has states. One of them is a goal state. The task is then the ``task completion'' type. If there are no apparent states, the task is more likely the ``cumulative gain'' type, to collect more gains at each step. All campaigns we mentioned above have both types.

 \section{Conclusion}\label{sec:conclusion}
IS systems serve as digital assistants through interacting with whom a human user acquires information and makes rational decisions or exciting discoveries. Plenty of research work has been generated on this topic and taxonomies have been developed to classify different types IS tasks. However, existing taxonomies are mostly proposed from the user's perspective and do not directly offer guidance to systems' algorithmic design. This may slow down the integration of the fast-advancing AI techniques into IS systems.   

In this article, we propose a new taxonomy for IS tasks, with the aim to reveal their underlying dimensions that can be easily computationalized. Our taxonomy identifies a few important dimensions for IS tasks. These dimensions include the number of search iterations, search goal's motivation types, and the types of procedures to reach the goals. We then review the state-of-the-art Information Seeking tasks and methods using the new taxonomy as a guidance.

The new taxonomy offers some unique angles to examine the existing literature and help scope active on-going research areas. For instance, our separation of ``task completion'' and ``cumulative gain'' tasks can provide scoping ideas when people working on task-oriented dialogue systems, conversational AI, without mixing together tasks having different natures. 

Another potential use of the taxonomy is when deciding what effective measures should be used to evaluate a method. If the task is ``task completion", the success rate would be a better effective metric than cumulative gain. On the other hand, if the task belongs to ``cumulative gain", then it is important to obtain a consistent and monotonically increasing gain curve. Moreover, if the task is mainly ``exploratory'', discovery, and fun-driven, the measure might be novel items discovered, amount of fun or time enjoying the search service. We can see these measures would be largely different but all about ``effectiveness", depending on what effective means to a task. Our taxonomy offers a clearer distinctions in the tasks and evaluations, expecting more research along the direction. 

Our article covers a broad range of related topics. It provides an overview to help navigate in the intersection of IR, IS, ML, and AI. By organizing the literature, knowing the known, we are then able to explore the unknown and less explored areas. For instance, open-ended RL and lifelong learning would be very interesting future directions for research in IS. 

Overall, our taxonomy puts a focus on revealing the computational dimensions of the IS tasks, which can help IR and ML practitioners to quickly decide what category a new project belongs to, and which family of computational methods should be used. It will help accelerate the initial exploration stage for a new project. Focusing on computational analysis of the IS tasks and methods, we hope this paper contribute to promote the system-side study of IS.

\section*{Acknowledgements}
The authors would like to thank the anonymous reviewers who greatly helped shaping our views. We thank Dr. Marti Hearst for her constructive feedback to an early version of this paper. We thank Dr. Jiyun Luo, Mr. Shiqi Liu, Dr. Ian Soboroff, Ms. Angela Yang, Mr. Limin Chen, and Dr. Xuchu Dong for their tremendous efforts during collaboration on this research. We thank the annotators from National Institute of Standards and Technology (NIST) who built the TREC DD evaluation datasets. We also thank Mr. Connor Lu for proof-reading the paper. This research was supported by NSF CAREER grant IIS-1453721 and DARPA Memex Program FA8750-14-2-0226. Any opinions, findings, conclusions, or recommendations expressed in this paper are of the authors, and do not necessarily reflect those of the sponsor.

\bibliographystyle{ACM-Reference-Format}
\bibliography{bibtex/reference_short}

\end{document}